\documentclass[runningheads]{llncs}
\usepackage[T1]{fontenc}
\usepackage[parfill]{parskip}
\usepackage{graphicx}
\usepackage{booktabs}
\usepackage{multirow}
\usepackage{amsmath,amsfonts,amssymb}
\usepackage{bbm}
\usepackage[hidelinks,colorlinks=true]{hyperref}
\usepackage{subcaption}
\usepackage{enumitem}
\usepackage{graphicx,verbatim}
\begin{document}
\title{RadAlign: Advancing Radiology Report Generation with Vision-Language Concept Alignment}
\titlerunning{RadAlign: Advancing Radiology Report Generation}

\author{Difei Gu\inst{1} \and Yunhe Gao\inst{1,2} \and Yang Zhou\inst{1} \and Mu Zhou\inst{1} \and Dimitris Metaxas\inst{1}}  
\authorrunning{Difei Gu \and Yunhe Gao \and Yang Zhou \and Mu Zhou \and Dimitris Metaxas}
\institute{Rutgers University \and Stanford University}

\maketitle              

\begin{abstract}

Medical image interpretation and report generation are essential for radiologists to identify and communicate observable findings of diseases. Major efforts in image-to-report generation require heavy language model training yet still suffer from producing reports with factual errors. In this study, we present RadAlign, demonstrating that a concept-based vision-language model can improve both predictive accuracy and report factual correctness without extensive language model training. Our key innovation is aligning visual features with medical diagnostic criteria in a shared representation space. Such alignment introduces core knowledge supervision and creates interpretable intermediate diagnosis results for LLMs to refine report generation. We also propose a cross-modal retrieval mechanism to provide additional clinical context of history cases for enhancing report generation accuracy. This unified approach achieves superior disease classification on MIMIC-CXR (average AUC: 0.885) and enables accurate report generation (GREEN score: 0.678 vs. SOTA: 0.634). RadAlign also demonstrates exceptional generalization capabilities, outperforming SOTA foundation and specialized models on the external OpenI dataset (AUC: 0.923 vs. 0.836). Code is available at \url{https://github.com/difeigu/RadAlign}.



\end{abstract}
\vspace{-0.5em}
\section{Introduction}
\vspace{-0.5em}

Medical image interpretation and report generation play a vital role in the clinical workflow that can directly impact disease characterization and patient care ~\cite{li2018hybrid}. For instance, chest radiograph interpretation~\cite{mcbee2018deep} remains as a critical task, where clinicians must recognize subtle abnormalities and translate precise disease classifications into detailed reports. Accomplishing this complex task requires systematic efforts to capture a detailed state of the disease and generate comprehensive, well-reasoned explanations of these clinical findings~\cite{reale2024vision}.

Major research on chest radiographic interpretation falls into classification models \cite{ye2020weakly,boecking2022making,asif2020classification} and image captioning approaches \cite{chen2022cross,jing2020show}. First, classification methods build on deep neural networks~\cite{he2016deep,asif2020classification,ye2020weakly} and vision transformers~\cite{dosovitskiy2020image,okolo2022ievit} to show diagnostic precision in detecting pneumonia, cardiomegaly, and pulmonary edema. However, these models operate as black boxes by predicting only disease labels without explaining the visual semantic features that led to their predictions. This lack of interpretability limits their utility in real-world clinics. Second, growing efforts have investigated image captioning approaches~\cite{jing2020show,chen2020generating,vinyals2015show,anderson2018bottom} for generating free-text radiology reports. Despite of their advance, these methods often require extensive language model training yet still suffer from hallucinations - generating incorrect or unreliable information misaligned with the actual image content or medical knowledge~\cite{ramesh2022improving,zhang2024data}.

 The alignment between visual content and language context \cite{gao2024aligning} is essential for developing human-level diagnostic reports. To illustrate, radiologists follow a structured process where they first assess specific diagnostic criteria and medical concepts (e.g., heart size, lung opacity, or pulmonary vessels) and then synthesize these observations with their medical knowledge to form detailed reports \cite{hodler2019diseases}. This clinical workflow motivates our development of \textbf{RadAlign}, a multi-modal framework that unifies the strength of predictive models with the reasoning capabilities of LLMs. Unlike prior approaches defining visual analysis and report generation as separate tasks~\cite{wang2023chatcad}, RadAlign purposely mirrors the radiologist's workflow on the concept-based image diagnosis. Our contributions are:

\begin{figure}[t]
    \centering
    \includegraphics[width=0.9\textwidth]{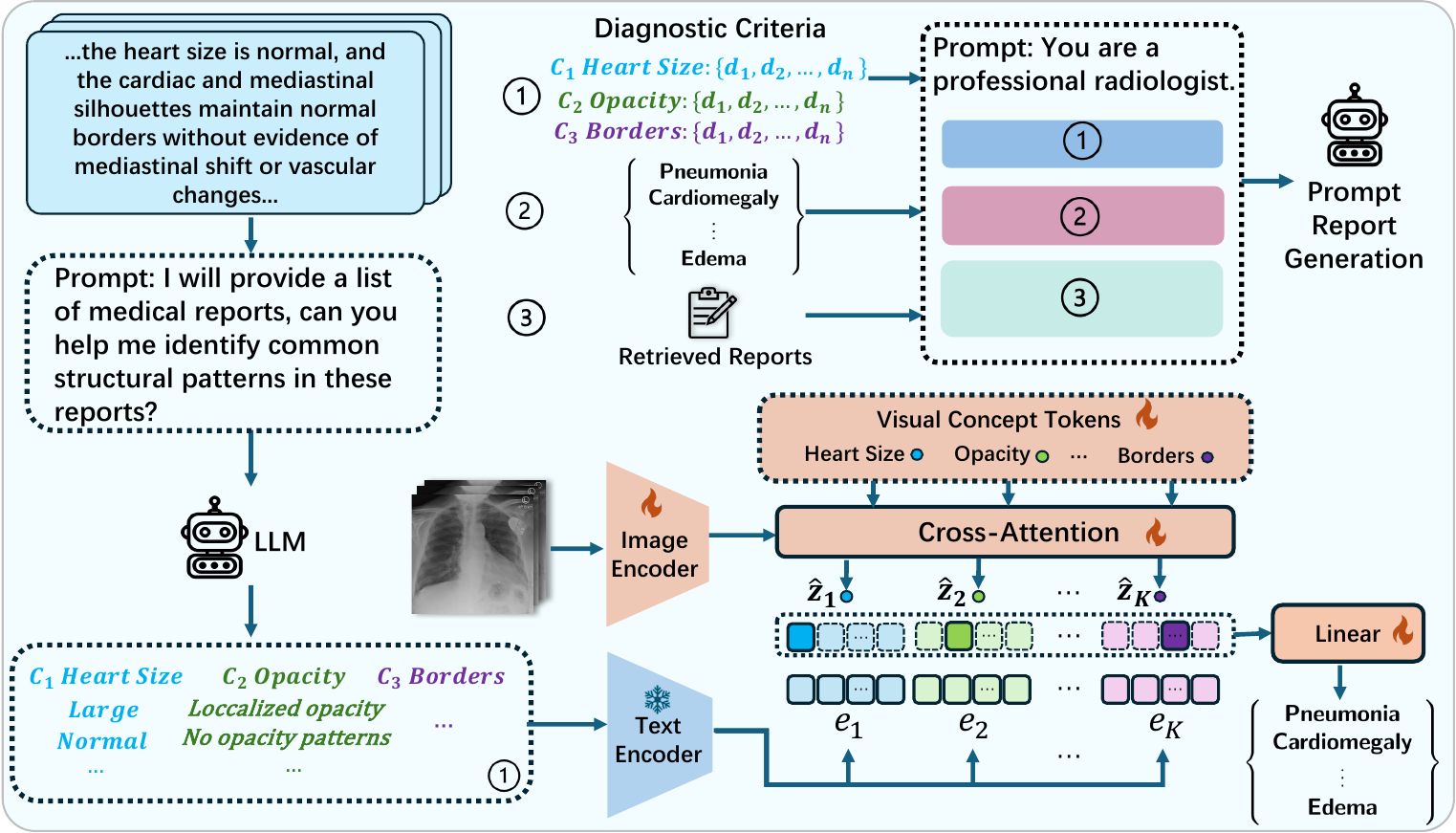}
    \caption{\textbf{Overview of the RadAlign framework.} Our unified VLM predicts three key components: \textbf{1. Diagnostic criteria} and associated concepts, formulated by an LLM based on historical reports, facilitate learning of image-concept alignment; \textbf{2. Disease prediction} is derived from this alignment, enabling an explainable classifier; \textbf{3. Augmented historical reports} are retrieved using learned visual concept tokens. These three components are combined and serve as knowledge-guided prompts for the LLM, ensuring a factually accurate report.}
    \label{fig:model}
    \vspace{-2em}
\end{figure}

\begin{itemize}[noitemsep,topsep=0pt,parsep=0pt,partopsep=0pt]
\item A unified framework that bridges the gap between classification accuracy and detailed reporting through the vision-language concept alignment.
\item A novel approach to medical report generation that mirrors radiologist workflow, combining visual feature recognition with LLM-based reasoning.
\item A cross-modal retrieval-augmented generation system that enhances report reliability by grounding predictions in similar historical cases.
\item Superior performance in classification (AUC: 0.885 on MIMIC-CXR, 0.923 on OpenI) and report generation benchmarks (GREEN score: 0.678 vs. SOTA: 0.634) with improved interpretability for clinical applications.
\end{itemize}

\vspace{-0.5em}
\section{Related Work}
\vspace{-0.5em}
\noindent\textbf{Vision-Language Models (VLMs)} seek to align visual and textual representations via contrastive learning and multimodal pre-training. General-purpose VLMs~\cite{radford2021learning,boecking2022making,li2023blip} trained on natural images often lack the specific medical knowledge required for disease understanding. Therefore, increased domain-specific adaptations have been explored including BioViL ~\cite{boecking2022making}, MedCLIP ~\cite{wang2022medclip} and MedKLIP~\cite{wu2023medklip}. These methods excel at learning joint representations to enable downstream tasks. They often focus on image-and-text matching in a pre-training framework, without explicitly considering patient information retrieval at inference time. In contrast, our effort emphasizes on learning domain-specific concepts, which serve as interpretable anchors for image analysis and textual reporting. This key procedure provides a transparent foundation for case retrieval and report generation, ensuring that the final report is clinically grounded. 

\noindent\textbf{Multimodel Caption Generation}
aims to integrate textual and visual information to improve the caption quality. Traditional approaches~\cite{chen2022cross,liu2021exploring,qin2022reinforced} employ neural networks to leverage both visual and textual features for generating coherent reports. However, these methods often function as black-box models, prioritizing the performance at the cost of interpretability. With the rise of large language models (LLMs), approaches like ChatCAD~\cite{wang2023chatcad} aim to incorporate reasoning by combining classification networks, segmentation models, report generation modules, and LLMs. It is clear that these independent models could introduce significant computational overhead and increased integration complexity. Inconsistencies can emerge when aligning outputs from different modules, potentially compromising the final report’s quality and reliability.

\vspace{-0.5em}
\section{Methodology}
\vspace{-0.5em}

\subsection{Domain Knowledge Query}
Inspired by how radiologists diagnose images, we first extract structured diagnostic criteria by mining expert-provided findings to create a foundation for concept-based diagnosis.
Let \(\mathcal{D}=\{(x, P, y)\}\) be our training set, where \(x\) is an image, \(P\) the ground-truth report findings, and \(y \in \mathcal{Y}\) the disease label among \(N\) classes.
We compile all findings \(\mathcal{P} = \{P_1, P_2, \ldots, P_{|\mathcal{D}|}\}\) and prompt an LLM to derive a set of \(K\) disentangled diagnostic criteria \(\{\mathcal{C}_i\}_{i=1}^K\).
For example, for chest X-rays, the criteria might include \texttt{Heart Size}, \texttt{Lung Opacity}, \texttt{Diaphragm Position}, etc.
We then query more detailed knowledge per criterion, grouping them by disease class as \(\mathcal{C}_i = \{C_i^1, C_i^2, \ldots, C_i^{n_i}\}\).
Each description indicates how that criterion manifests for each disease (e.g., \texttt{Heart Size} changes for \emph{Cardiomegaly} but not for \emph{Pneumonia}).
Lastly, we build a mapping \(f_m: \mathcal{C}\rightarrow \mathcal{Y}\) to link each concept description to one or more disease classes.This structured knowledge extraction provides crucial semantic anchors for our model to learn clinically relevant patterns.

\vspace{-0.7em}
\subsection{Visual Concept Fine-grained Alignment}

This component aims to discover and recognize specific visual features based on the structured diagnostic criteria, enabling the model to "see" like a radiologist. Using a pretrained vision-language model, we encode the textual criteria \(\{\mathbf{e}_i\}_{i=1}^K\) via its text encoder \(\mathcal{T}\).
Each \(\mathbf{e}_i \in \mathbb{R}^{n_i \times d}\) anchors the expert-derived concepts in embedding space.
Meanwhile, we introduce \(K\) learnable visual concept tokens \(\mathbf{z} \in \mathbb{R}^{K \times d}\) in the visual encoder \(\mathcal{V}\).
Given an image \(x\), we extract features \(\mathcal{V}(x)\) and use cross-attention to obtain:
\begin{equation}
    \hat{\mathbf{z}} = \text{cross-attention}(\mathbf{z}, \mathcal{V}(x), \mathcal{V}(x)),
\end{equation}
where \(\mathbf{z}\) acts as the query.
Each of the \(K\) tokens is encouraged to focus on the visual features pertinent to its corresponding criterion. To align visual and textual embeddings, we employ a domain-specific contrastive loss:
\begin{equation}
    \mathcal{L}_{anchor}^{i}(\hat{\mathbf{z}}_i, \mathbf{e}_i) 
    = -\log \frac{\exp\bigl(\text{sim}(\hat{\mathbf{z}}_i, \mathbf{e}_i^{\text{positive}})/\tau\bigr)}{\sum_{j=1}^{n_i}\exp\bigl(\text{sim}(\hat{\mathbf{z}}_i, \mathbf{e}_i^j)/\tau\bigr)},
\end{equation}
where \(\hat{\mathbf{z}}_i\) and \(\mathbf{e}_i\) are matched concept embeddings, \(\tau\) is a temperature parameter, and \(\text{sim}\) denotes dot-product similarity. This alignment process teaches the model to recognize clinically relevant patterns in radiographs, mimicking how radiologists diagnose with criteria.

\vspace{-0.7em}
\subsection{Knowledge-guided Prompting}

We propose a novel approach that leverages the diagnostic power of our aligned concept model without requiring extensive language model training. Our key insight is that the well-aligned concept-based vision-language model already contains sufficient diagnostic information for accurate report generation.
Our vision-language model produces visual concept tokens $\mathbf{\hat{z}}_i$ aligned with diagnostic criteria anchors $\mathbf{e}_i$. We construct an explainable classifier using the similarity:
\begin{equation}
\hat{y} = W (\text{concat}(sim(\hat{\mathbf{z}}_1, \mathbf{e}_1), \dots, sim(\hat{\mathbf{z}}_K, \mathbf{e}_K)))^{\intercal},
\end{equation}
where $W$ represents the significance of each criterion's contribution toward classification. The total loss function combines cross-entropy for disease classification with the average contrastive loss:

\begin{equation}
\mathcal{L}_{total}
= \mathcal{L}_{\text{ce}}(\hat{y}, y)
+ \tfrac{1}{K} \sum_{i=1}^{K}
\mathcal{L}_{anchor}^i(\hat{\mathbf{z}}_i, \mathbf{e}_i).
\end{equation}
To generate the report, we directly prompt the LLM with both the recognized criteria from our aligned model and the final classification prediction. Since the vision-language model is aligned in terms of diagnostic criteria concepts, it already contains the detailed findings necessary for accurate reporting. The LLM's role is primarily to reform these findings into a coherent, well-structured report rather than making diagnostic decisions. This approach uniquely combines the accuracy of our predictive model with the language capabilities of LLMs while significantly reducing hallucinations, as the factual diagnostic content is already ensured by the vision-language alignment.

\vspace{-0.7em}
\subsection{Image-based Report Retrieval Augmentation (RAG)}
While our aligned concept model provides accurate diagnostic findings, we recognize that general-purpose LLMs are not specifically trained for medical reporting. They require clinical context to understand appropriate radiology writing styles and terminology. Our image-based RAG system addresses this by providing relevant clinical examples that help the LLM reason more effectively for medical reporting.
We construct a report database of training images:
\begin{equation}
    \mathcal{Q} = \{(\hat{z}_i, P_i)\}_{i=1}^{|D|}
\end{equation}
Where $(\hat{z}_i, P_i)$ is a key-value pair of visual concept tokens and corresponding reports. We precompute and store the visual concept tokens to minimize inference overhead. For each image $x \not\in D$, we retrieve the most similar TopK cases:
\begin{equation}
    \mathcal{P}_{\text{retrieve}} = \mathcal{Q}(\hat{z}_i, \hat{z}_i \in \text{TopK}_{\hat{z}\in\{\hat{z}_1, \hat{z}_2, ..., \hat{z}_{|D|}\}}sim(\hat{z}_i, \mathbf{\hat{z}_{x}}))
\end{equation}
Where $\mathbf{\hat{z}_{x}}$ is the predicted concept token of any testing image. This retrieval mechanism grounds the LLM's output in validated clinical examples, helping maintain professional terminology and reporting conventions. By providing similar cases with confirmed diagnoses, we enable the LLM to better contextualize the aligned visual concepts into a clinically appropriate report. The classification results, concept findings, and retrieved reports are then incorporated into a unified prompt for the LLM, leveraging its editing capabilities while ensuring medical accuracy and relevance.

\vspace{-0.5em}
\section{Experiment and Results}
\vspace{-0.2em}
\subsection{Experimental Setup}

\textbf{Dataset.} We use \textbf{MIMIC-CXR} \cite{johnson2019mimic} for a comprehensive training and evaluation. The MIMIC-CXR dataset contains 377,100 chest X-ray images and corresponding radiology reports including findings, impressions, and patient history. We use five common classes including Atelectasis (AT), Cardiomegaly (CM), Consolidation (CD), Edema (ED) and Pleural Effusion (PE). To highlight RadAlign's out-of-domain generalization, we further evaluate on \textbf{IU X-ray (OpenI)}\cite{demner2016preparing}, which is unseen by all comparison methods. IU X-ray contains 7,470 chest X-ray images with corresponding reports from 3,955 patients. 

\textbf{Baselines.} We evaluate our model against SOTA baselines for both disease classification and report generation tasks. For disease classification, we compare with: zero-shot foundation models, like CLIP \cite{radford2021learning}, BiomedicalCLIP \cite{zhang2023biomedclip}, and BioViL \cite{boecking2022making}. We also compare with domain-specific models that are trained on the MIMIC-CXR dataset, including PCAM \cite{ye2020weakly}; ChatCAD \cite{wang2023chatcad}, a multi-model integration LLM framework;  LABO \cite{yang2023language}, an explainable VLM with concept bottleneck. For report generation, we compare against R2GenCMN \cite{chen2022cross}, a cross-modal memory network for visual-textual integration, and ChatCAD \cite{wang2023chatcad}.

\textbf{Implementation Details.} We prompt GPT-4 to query the diagnostic criteria. For our vision backbone, we conduct experiments using both ImageNet pretrained Resnet-50 weights and the BioViL CLIP Resnet-50 weights~\cite{boecking2022making}. During training, we only optimize the visual encoder, visual concept tokens, and the final linear layer with AdamW optimizer for 40 epoches, using a decreasing learning rate of [1e-3, 1e-4], while keeping the text encoder fixed. All experiments are conducted using PyTorch with Nvidia RTX 8000 GPUs.

\begin{table}[tb]
    \centering
    \scriptsize
    \caption{\small \textbf{Left}: Report generation comparison. \textbf{Right}: RadAlign with different LLMs.}
    \label{tab:report_generation_main_result}
    \begin{minipage}{0.47\textwidth}
        \centering
        \setlength{\tabcolsep}{2.5mm}
        \begin{tabular}{c|c|c}
            \toprule
            Model               & LLM     & GREEN $\uparrow$\\
            \midrule
            R2GenCMN            & -       & 0.634\\
            ChatCAD             & 4o-mini & 0.633\\
            ChatCAD             & 4o      & 0.634\\
            \midrule
            $\text{RadAlign}^{\dagger}$ & 4o-mini & 0.629\\
            $\text{RadAlign}^{\dagger\dagger}$ & 4o-mini & 0.648\\
            $\text{RadAlign}^{\dagger\dagger}$     & 4o      & \textbf{0.678}\\
            \bottomrule
        \end{tabular}
        \\\quad\quad\raggedright $^{\dagger}$ Initialized with ImageNet weights. \\
        \quad\quad\raggedright $^{\dagger\dagger}$ Initialized with BioViL weights.
    \end{minipage}
    \hfill
    \begin{minipage}{0.47\textwidth}
        \centering
        \setlength{\tabcolsep}{2.5mm}
        \begin{tabular}{c|c}
            \toprule
            RadAlign + LLM                & GREEN $\uparrow$\\
            \midrule
            ChatGPT 3.5-Turbo  & 0.648\\
            ChatGPT 4o-mini    & 0.646\\
            ChatGPT 4o         & 0.678\\
            Claude 3.5-Sonnet  & 0.658\\
            Llama 3.1          & 0.695\\
            \bottomrule
        \end{tabular}
    \end{minipage}
    \vspace{-2em}
\end{table}

\vspace{-0.7em}
\subsection{Results}
\vspace{-0.2em}
\textbf{Report Generation Comparison.} We evaluate report generation quality using GREEN Score~\cite{ostmeier2024green}, a metric specifically designed for medical report assessment that leverages LLM-based reasoning to identify clinically significant errors. Unlike traditional metrics such as BLEU \cite{papineni2002bleu}, ROUGE \cite{lin-2004-rouge}, and BERTScore \cite{zhang2019bertscore} that only measure surface-level text similarity without considering factual correctness, GREEN focuses on accurately distinguishing between presence and absence of conditions and offers both quantitative scores and interpretable explanations that align well with expert judgment. For implementation details and more discussion, we refer readers to the original paper~\cite{ostmeier2024green}.

As shown in Table \ref{tab:report_generation_main_result} (left), RadAlign achieves superior performance with a GREEN score of 0.678 using GPT-4o, substantially outperforming the baseline methods (0.634). Notably, we observe different scaling behaviors between methods: ChatCAD shows negligible improvement when upgrading from GPT-4o mini to GPT-4o, while RadAlign shows significant performance gains (0.648 to 0.678). This differential scaling highlights how RadAlign’s unified vision-language alignment effectively leverages enhanced LLM reasoning capabilities based on recognized medical concepts, while ChatCAD’s multi-model pipeline lacks alignment, introducing inconsistencies that limit the benefits of more powerful LLMs.

\textbf{Classification Accuracy Comparison.} Table \ref{tab:combined_classification_scores} presents  disease classification results in terms of F1 score and AUC. On MIMIC-CXR, RadAlign achieves the leading classification performance with an average F1 score of 0.633 and AUC of 0.885, outperforming both foundation models like BiomedCLIP, BioVil, and specialized methods like ChatCAD and PCAM. More impressively, when evaluated on the unseen OpenI dataset, RadAlign maintains strong performance with an average F1 score of 0.652 and AUC of 0.923, demonstrating excellent generalization capability and robustness to domain shifts.

\begin{table}[tb]
    \centering
    \scriptsize
    \caption{Classification results for different methods on F1 and AUC}
    \label{tab:combined_classification_scores}
    \begin{subtable}[t]{\textwidth}
        \centering
        \caption{MIMIC-CXR}
        \setlength{\tabcolsep}{0.7mm}
        \begin{tabular}{l *{6}{cc}}
            \toprule
            Model & \multicolumn{2}{c}{AT} & \multicolumn{2}{c}{CM} & \multicolumn{2}{c}{CD} & \multicolumn{2}{c}{ED} & \multicolumn{2}{c}{PE} & \multicolumn{2}{c}{Average} \\
            \cmidrule(lr){2-3} \cmidrule(lr){4-5} \cmidrule(lr){6-7} \cmidrule(lr){8-9} \cmidrule(lr){10-11} \cmidrule(lr){12-13}
             & F1 & AUC & F1 & AUC & F1 & AUC & F1 & AUC & F1 & AUC & F1 & AUC \\
            \midrule
            CLIP               & 0.200 & 0.507 & 0.200 & 0.540 & 0.000 & 0.497 & 0.060 & 0.498 & 0.200 & 0.500 & 0.132 & 0.508 \\
            BiomedCLIP         & 0.180 & 0.547 & 0.113 & 0.526 & 0.157 & 0.584 & 0.166 & 0.572 & 0.365 & 0.614 & 0.196 & 0.569 \\
            BioViL             & 0.388 & 0.705 & 0.431 & 0.715 & 0.165 & 0.806 & 0.329 & 0.783 & 0.582 & 0.769 & 0.379 & 0.756 \\
            $\text{PCAM}^{*}$             & 0.618 & 0.838 & 0.628 & \textbf{0.876} & 0.432 & 0.787 & 0.514 & 0.868 & 0.755 & 0.937 & 0.589 & 0.861 \\
            ChatCAD            & 0.311 & 0.542 & 0.523 & 0.650 & 0.527 & 0.724 & \textbf{0.641} & 0.662 & 0.764 & 0.838 & 0.553 & 0.683 \\
            LABO               & 0.583 & 0.753 & 0.607 & 0.768 & 0.462 & 0.747 & 0.556 & 0.820 & 0.714 & 0.847 & 0.584 & 0.787 \\\hline
            $\text{RadAlign}^{\dagger}$   & 0.628 & 0.841 & 0.650 & 0.873 & \textbf{0.490} & \textbf{0.824} & 0.616 & 0.916 & 0.779 & \textbf{0.956} & \textbf{0.633} & 0.882 \\
            $\text{RadAlign}^{\dagger\dagger}$ & \textbf{0.634} & \textbf{0.853} & \textbf{0.653} & 0.873 & 0.473 & \textbf{0.824} & 0.580 & \textbf{0.924} & \textbf{0.820} & 0.954 & 0.632 & \textbf{0.885} \\
            \bottomrule
        \end{tabular}
        \vspace{0em}
        
        $^{*}$ MIMIC-CXR finetuned. $^{\dagger}$ Initialized with ImageNet weights. $^{\dagger\dagger}$ Initialized with BioViL weights.
    \end{subtable}
    
    \begin{subtable}[t]{\textwidth}
        \centering
        \caption{IU X-ray (OpenI)}
        \setlength{\tabcolsep}{0.7mm}
        \begin{tabular}{l *{6}{cc}}
            \toprule
            Model & \multicolumn{2}{c}{AT} & \multicolumn{2}{c}{CM} & \multicolumn{2}{c}{CD} & \multicolumn{2}{c}{ED} & \multicolumn{2}{c}{PE} & \multicolumn{2}{c}{Average} \\
            \cmidrule(lr){2-3} \cmidrule(lr){4-5} \cmidrule(lr){6-7} \cmidrule(lr){8-9} \cmidrule(lr){10-11} \cmidrule(lr){12-13}
             & F1 & AUC & F1 & AUC & F1 & AUC & F1 & AUC & F1 & AUC & F1 & AUC \\
            \midrule
            CLIP               & 0.272 & 0.517 & 0.249 & 0.559 & 0.000 & 0.499 & 0.046 & 0.502 & 0.101 & 0.500 & 0.134 & 0.515 \\
            BiomedCLIP         & 0.100 & 0.520 & 0.160 & 0.542 & 0.200 & 0.618 & 0.118 & 0.550 & 0.348 & 0.610 & 0.185 & 0.568 \\
            BioViL             & 0.272 & 0.517 & 0.249 & 0.559 & 0.000 & 0.499 & 0.046 & 0.502 & 0.101 & 0.500 & 0.134 & 0.515 \\
            $\text{PCAM}^{*}$      & 0.540 & 0.569 & 0.715 & 0.862 & \textbf{0.607} & \textbf{0.978} & 0.505 & 0.809 & \textbf{0.786} & 0.961 & 0.630 & 0.836 \\\hline
            $\text{RadAlign}^{\dagger\dagger}$   & \textbf{0.695} & \textbf{0.851} & \textbf{0.737} & \textbf{0.913} & 0.563 & 0.952 & \textbf{0.618} & \textbf{0.934} & 0.648 & \textbf{0.963} & \textbf{0.652} & \textbf{0.923} \\
            \bottomrule
        \end{tabular}
        \vspace{0em}
        
        $^{*}$ MIMIC-CXR finetuned. $^{\dagger\dagger}$ Initialized with BioViL weights.
    \end{subtable}
    \vspace{-2em}
\end{table}

\begin{figure}[tb]
    \centering
    
    \begin{minipage}{\textwidth}
        \centering
        \includegraphics[width=0.8\textwidth]{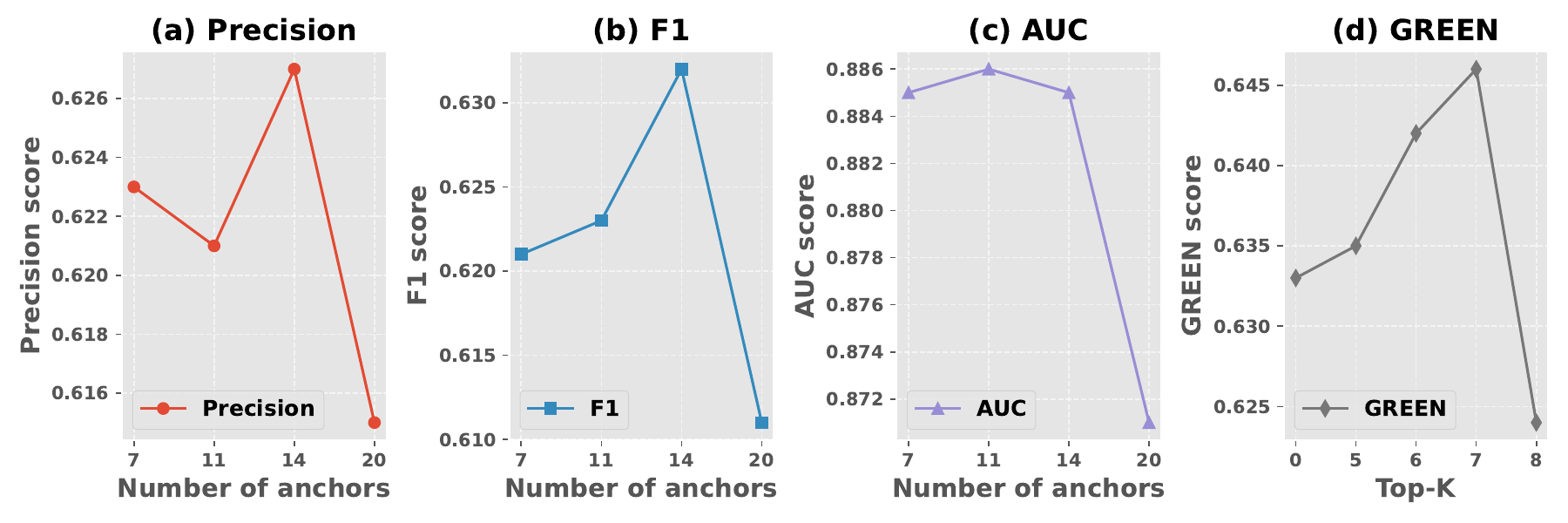}
    \end{minipage}
    
    \vspace{0em} 
    
    \begin{minipage}{\textwidth}
        \centering
        \includegraphics[width=0.75\textwidth]{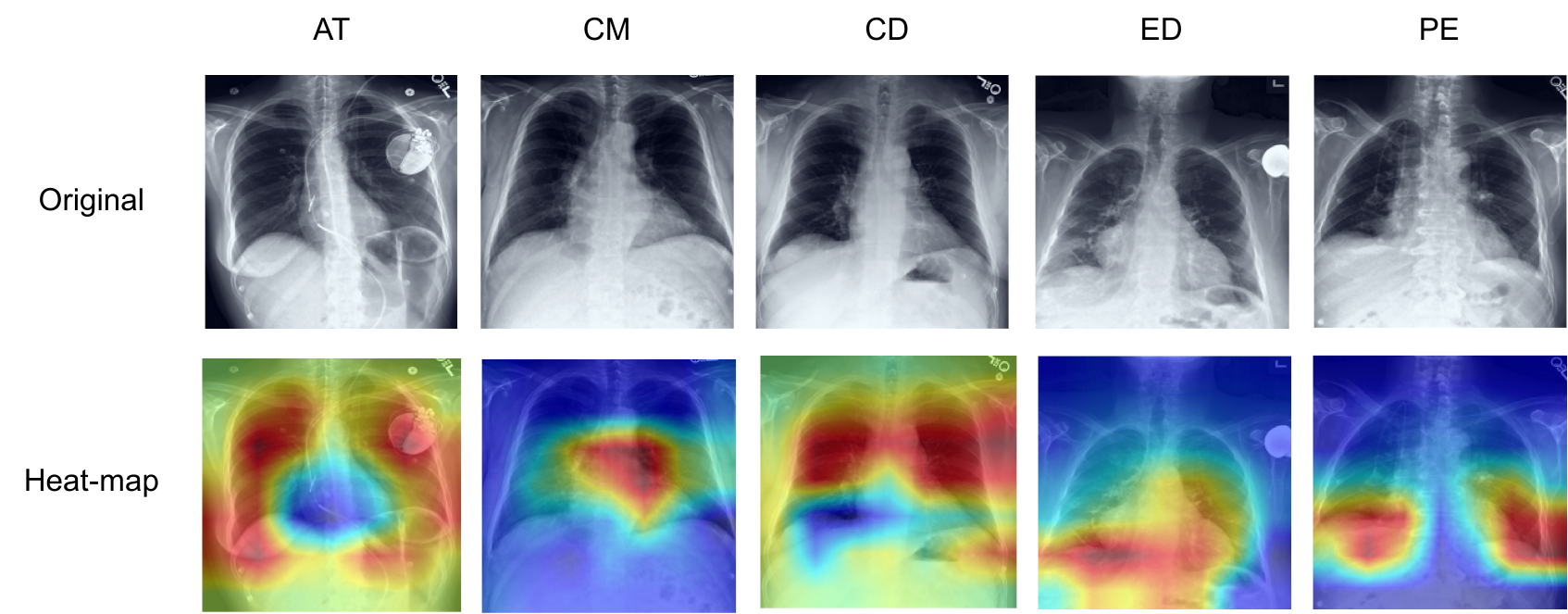}
    \end{minipage}
    
    \caption{\textbf{Top: Ablation studies}.  (a-c) show classification performance of different number of concept anchors. (d) illustrates GREEN-score performance when varying the number of retrieved similar reports $K$. \textbf{Bottom: Visualization}. Attention map of concept tokens for different disease classes (AT, CM, CD, ED, PE), with warmer colors indicating higher attention scores.}
    \label{fig:combined}
    \vspace{-2em}
\end{figure}

\textbf{Evaluation with different LLMs.} 
We evaluated RadAlign with various large language models to assess its compatibility and generalizability,as shown in Table \ref{tab:report_generation_main_result} (right). All tested LLMs, including ChatGPT (3.5, 4o-mini, 4o), Claude 3.5-Sonnet, and Llama 3.1 7B, achieved higher GREEN scores (0.646–0.695) than the previous baseline of 0.634, showing that our visual concept alignment approach is robust across different LLM architectures. More advanced models generally performed better (e.g., ChatGPT improved from 0.648 to 0.678).

\textbf{Ablation Studies.}
Our experiments identified optimal parameters for RadAlign: performance peaked with 14 concept anchors across all metrics (Fig. \ref{fig:combined}, top a-c), as additional anchors introduced noise rather than meaningful features. For report retrieval, K=7 similar reports yielded the highest GREEN-Score (Fig. \ref{fig:combined}, top d), balancing sufficient semantic guidance without introducing misleading information from less relevant cases.

\textbf{Concept Interpretation.}
RadAlign enables transparent interpretation of its decision-making process through visualization of concept token attention weights, displaying disease-specific localization patterns that align with clinical expertise. In Fig. \ref{fig:combined} bottom, the attention heatmaps highlight anatomically-relevant regions for each condition. For example, for Atlectasis (AT), the heatmap highlights specific areas around the edge of the lung fields that are indicative of abnormalities, while for Cardiomegaly (CM), attention is drawn to distinct location at the heart region. These visualizations validate that our concept tokens can capture clinically meaningful features to assess the model's reasoning process.

\vspace{-0.5em}
\section{Discussion and Conclusion}
\vspace{-0.2em}

We introduce RadAlign, a novel framework that aligns visual features with medical concepts using a specialized Vision-Language Model. Unlike conventional methods that rely on extensive language model training or basic LLM prompting, RadAlign leverages a robust, concept-driven alignment strategy to map image features to diagnostic criteria. RadAlign achieves superior disease classification with an AUC of 0.885 on MIMIC-CXR and 0.923 on OpenI, while generating high-quality reports with a GREEN score of 0.678, outperforming state-of-the-art baselines. By integrating retrieval-augmented generation, RadAlign enhances factual accuracy and interpretability, drawing on similar historical cases to reduce hallucination. Notably, RadAlign bypasses extensive language model training, offering an efficient solution for clinical applications. Its concept-driven design ensures transparency by mirroring radiologists’ diagnostic workflows.

\begin{credits}
\subsubsection{\ackname} This research has been partially funded by research grants to D. Metaxas through NSF: 2310966, 2235405, 2212301, 2003874, 1951890 and NIH 2R01HL127661.

\subsubsection{\discintname}
The authors have no competing interests to declare
that are relevant to the content of this article. 
\end{credits}
%
%
%
\bibliographystyle{splncs04}
\bibliography{reference.bib}
%




\end{document}